# A Deep Convolutional Auto-Encoder with Pooling - Unpooling Layers in Caffe

Volodymyr Turchenko, Eric Chalmers, Artur Luczak

Canadian Centre for Behavioural Neuroscience
Department of Neuroscience, University of Lethbridge
4401 University Drive, Lethbridge, AB, T1K 3M4, Canada
{vtu, eric.chalmers, luczak}@uleth.ca

*Abstract* – This paper presents the development of several models of a deep convolutional auto-encoder in the Caffe deep learning framework and their experimental evaluation on the example of MNIST dataset. We have created five models of a convolutional auto-encoder which differ architecturally by the presence or absence of pooling and unpooling layers in the auto-encoder's encoder and decoder parts. Our results show that the developed models provide very good results in dimensionality reduction and unsupervised clustering tasks, and small classification errors when we used the learned internal code as an input of a supervised linear classifier and multi-layer perceptron. The best results were provided by a model where the encoder part contains convolutional and pooling layers, followed by an analogous decoder part with deconvolution and unpooling layers without the use of switch variables in the decoder part. The paper also discusses practical details of the creation of a deep convolutional auto-encoder in the very popular Caffe deep learning framework. We believe that our approach and results presented in this paper could help other researchers to build efficient deep neural network architectures in the future.

*Keywords* – Deep convolutional auto-encoder, machine learning, neural networks, dimensionality reduction, unsupervised clustering.

## 1. Introduction

An auto-encoder (AE) model is based on an encoder-decoder paradigm, where an encoder first transforms an input into a typically lower-dimensional representation, and a decoder is tuned to reconstruct the initial input from this representation through the minimization of a cost function [1-4]. An AE is trained in unsupervised fashion which allows extracting generally useful features from unlabeled data. AEs and unsupervised learning methods have been widely used in many scientific and industrial applications, solving tasks like network pre-training, feature extraction, dimensionality reduction, and clustering. A classic or shallow AE has only one hidden layer which is a lower-dimensional representation of the input. In the last decade, the revolutionary success of deep neural network (NN) architectures has shown that deep AEs with many hidden layers in the encoder and decoder parts are the state-of-the-art models in unsupervised learning. In comparison with a shallow AE, when the number of trainable parameters is the same, a deep AE can reproduce the input with lower reconstruction error [5]. A deep AE can extract hierarchical features by its hidden layers and, therefore, substantially improve the quality of solving specific task. One of the variations of a deep AE [5] is a deep convolutional auto-encoder (CAE) which, instead of fully-connected layers, contains convolutional layers in the encoder part and deconvolution layers in the decoder part. Deep CAEs may be better suited to image processing tasks because they fully utilize the properties of convolutional neural networks (CNNs), which have been proven to provide better results on noisy, shifted (translated) and corrupted image data [6].

Modern deep learning frameworks, i.e. ConvNet2 [7], Theano with lightweight extensions Lasagne and Keras [8-10], Torch7 [11], Caffe [12], TensorFlow [13] and others, have become very popular tools in deep learning research since they provide fast deployment of state-of-the-art deep learning models along with state-of-the-art training algorithms (Stochastic Gradient Descent, AdaDelta, etc.) allowing rapid research progress and emerging commercial applications. Moreover, these frameworks implement many state-of-the-art approaches to network initialization, parametrization and regularization, as well as state-of-the-art example models. Besides many outstanding features, we have chosen the Caffe deep learning framework [12] mainly for two reasons: (i) a description of a deep NN is pretty straightforward, it is just a text file describing the layers and (ii) Caffe has a Matlab wrapper, which is very convenient and allows getting Caffe results directly into a Matlab workspace for their further processing (visualization, etc.) [12].

The goal of this paper is to present the practical implementation of several CAE models in the Caffe deep learning framework, as well as experimental results on solving an unsupervised clustering task using the MNIST dataset. This study is an extended version of our paper published in arXiv [14]. All developed Caffe *.prototxt* files to reproduce our models along with Matlab-based visualization scripts are included in supplementary materials. The paper is organized as follows: Section 2 describes relevant related work, Section 3 presents the developed models along with practical rules of thumb we have used to build the models, Section 4 presents the experimental results, Section 5 discusses the observations we have received in our experiments and Section 6 contains conclusions.

## 2. Related Work

Research on AE models was accelerated just after a breakthrough in artificial NNs connected with the success of the back propagation training algorithm in 1986 [1-4] and successfully continued a decade ago [5, 15]. There are many studies presenting advanced regularization and parametrization techniques for non-convolutional AEs including deep [5], denoising [16, 17], transforming [18], contractive [17, 19], k-sparse [20], variational [21], importance-weighted [22] and adversarial [23] AEs.

The work of Ranzato et al. [15] is one of the first studies which uses convolutional layers for unsupervised learning of sparse hierarchical features. Their model consists of two levels of CAE; each CAE has convolutional and max-pooling layers in the encoder part and upsampling and full convolutional layers in the decoder part. The model is trained independently using a greedy layer-wise approach, and the output of the first CAE level serves as an input to the second level. The authors suggested that if an AE may learn the identity function in a trivial way and produce uninteresting features, its hidden layer describes a code which can be overcomplete. Making this code sparse is a way to overcome this disadvantage. Lee et al. [24] and Norouzi et al. [25] have researched unsupervised learning of hierarchical features using a stack of convolutional Restricted Boltzmann Machines (RBM) and a greedy layer-wise training approach. The fully-connected operations were substituted by convolutional operations, and the probabilistic max-pooling was introduced in [24]. Deterministic max-pooling was used in [25]. Both Lee and Norouzi argue that convolutional RBMs have increased overcompleteness of the learned features and suggest adding sparsity to the hidden features. Masci et al. [26] have investigated shallow and deep (stacked) CAEs for hierarchical feature extraction, trained by a greedy layer-wise approach. Valid convolutional layers with and without max-pooling are used in the encoder part and full convolutional layers are used in the decoder part. The authors have stated that the use of max-pooling layers is an elegant way to provide the architecture with enough sparsity, and no additional regularization parameters are needed.

The closest work to our results is a recent paper by Zhao et al. [27] entitled "Stacked What-Where Auto-Encoders" (SWWAE). Their architecture integrates discriminative and generative pathways and provides a unified approach to supervised, semi-supervised and unsupervised learning. Within the unsupervised part, their SWWAE consists of several convolutional and max-pooling layers followed by one fully-connected layer in the encoder part and, inversely, one fully-connected, unpooling and deconvolution layers in the decoder part. Their SWWAE is symmetric in the encoder and decoder parts. The terms "what" and "where" correspond to pooling and appropriate unpooling operations which were proposed in [28-29]. The output of a max-pooling layer is the "what" variable (it is a max-pooled feature), which is fed to the next layer in the encoder part. The complementary "where" variables are the max-pooling "switch" positions. The "what" variables inform the next layer about the content with incomplete information about position, while the "where" variables inform the corresponding feed-back unpooling layer in the decoder part about where these max-pooled (dominant) features should be reconstructed in the unpooled feature map. In addition to a standard L2 (Euclidean) reconstruction cost function at the input level, the authors have proposed to use an L2 middle reconstruction cost function between the corresponding hidden layers in the encoder and decoder parts to provide better model training. Similarly to some other solutions, the authors have used a dropout layer [30] added to the fully-connected layers and an L1 sparsity penalty on hidden layers as a regularization technique.

The high quality of the extracted latent hierarchical features in the existing solutions analyzed above has been confirmed by the state-of-the-art classification results obtained by different classifiers trained in a supervised way. The extracted features were used as inputs for these classifiers. However, [23] was the only paper to provide a visualization of the extracted features in a two-dimensional space. We believe that such visuals might be considered as an inherent addition to the results presented in all studies above.

The use of a pooling layer is still the most controversial question in the theory and practice of CNNs [31]. The state-of-the-art approach in building supervised convolutional models is to have a max-pooling layer which computes the maximum activation of the units in a small region of the previous convolutional layer. Encoding the result of convolution operation with max-pooling allows higher-layer representations to be invariant to small translations of the input and reduces computational cost [24]. Scherer et al. [32] have shown that a max-pooling operation is considerably better at capturing invariances in image data, compared to subsampling operation. In unsupervised convolutional models Masci et al. [26] have shown that a CAE without max-pooling layers learns trivial solutions and interesting and biology plausible filters only emerge once a CAE is trained with a max-pooling layer. Zhao et al. [27] have proven that their SWWAE with max-pooling – unpooling layers provides much better quality of image reconstruction than max-pooling – unsampling layers. Controversially, recent findings by Springenberg et al. [33] have proven that a max-pooling operation can simply be replaced by a convolutional operation with increased stride without decreasing accuracy on several image recognition benchmarks. In our previous paper we have presented a CAE without pooling – unpooling layers which provided acceptable quality of the dimensionality reduction and unsupervised clustering tasks [14]. Therefore we have included the pooling - unpooling layers in our study below aiming to find which model, with or without pooling – unpooling layers, will be better.

Taking into account that a deep unsupervised model is a complex model from a training perspective, the question about a training approach is very important. We have chosen Caffe to use in our research, which implements the state-of-the-art training approach called "top-down" [29]; other terms are "jointly trained multiple layers" [34] or "jointly trained models" [27]. A top-down approach implies efficient training of all hidden layers of a model with respect to the input, while a greedy layer-wise training approach [15, 26] specifies that each layer receives its input from the latent representation of the layer below and trains independently. Zeiler et al. argued that the major drawback to a greedy layer-wise approach "is that the image pixels are discarded after the first layer, thus higher layers of the model have an increasingly diluted connection to the input. This makes learning fragile and impractical for models beyond a few layers" [29]. Moreover, Zeiler et al. have proven their conclusions by the comparison of top-down and greedy layer-wise approaches showing significantly better performance of the former on the Caltech-101 database. The advantage of a top-down training approach over a greedy layer-wise has been proven in the SWWAE study too [27].

Also there are several practical solutions/attempts to develop a CAE model on different platforms: shallow CAE [35] and convolutional RBM [36] in Matlab, deep CAE in Theano/Lasagne [37], Theano/Keras [38], Torch7 [39-40] and Neon [41].

## 3. Model description

In general, Caffe models are end-to-end machine learning systems. A typical network begins with a data layer which loads data from a disk and ends with one or several loss layers which specify a goal of learning (also known as an error, cost, objective or loss function). Our CAE models include convolutional, pooling, fully-connected, deconvolution, unpooling and loss layers. For brevity we do not include a description of the fully connected layer, which is pretty well-known.

The convolutional/deconvolution layer followed by an activation function is described by the expression [26]

$$h^k = f(\sum_{l \in L} x^l \otimes w^k + b^k),$$

where $h^k$ is the latent representation of the $k$-th feature map of the current layer, $f$ is an activation function (normally non-linear), $x^l$ is $l$-th feature map of the group of feature maps $L$ of the previous layer or $l$-th-channel of input image with total $L$ channels in a case of the first convolutional layer, $\otimes$ denotes the 2D convolution operation, $w^k$ and $b^k$ are the weights (filters) and the bias of the $k$-th feature map of the current layer respectively. If $x^l$ is an image or a feature map with size $m \times m$ and the filters are $n \times n$, a convolutional layer performs 'valid convolution' and the size of the output feature map $(m - n + 1) \times (m - n + 1)$ is decreasing. A deconvolution layer performs 'full convolution' and the size of the output feature map $(m + n - 1) \times (m + n - 1)$ is increasing [26, 28]. This is how convolutional layers provide encoding of an input image by decreasing the output feature maps from layer to layer in the encoder part and, inversely, how deconvolution layers provide reconstruction of the input image by the increasing the output feature maps from layer to layer in the decoder part.

A max-pooling layer pools features by taking the maximum activity within input feature maps (outputs of the previous convolutional layer) and produces (i) its output feature map with reduced size according to the size of pooling kernel and (ii) supplemental switch variables (switches) which describe the position of these max-pooled features [28-29]. The unpooling layer restores the max-pooled feature (what) either (i) into the correct place, specified by the switches (where), or, (ii) into some specific place within the unpooled output feature map. Fig. 1 illustrates convolution – deconvolution and pooling – unpooling operations. We have used a Caffe implementation of unpooling layer provided by Noh et al. [42], which was successfully applied for semantic segmentation task [43].

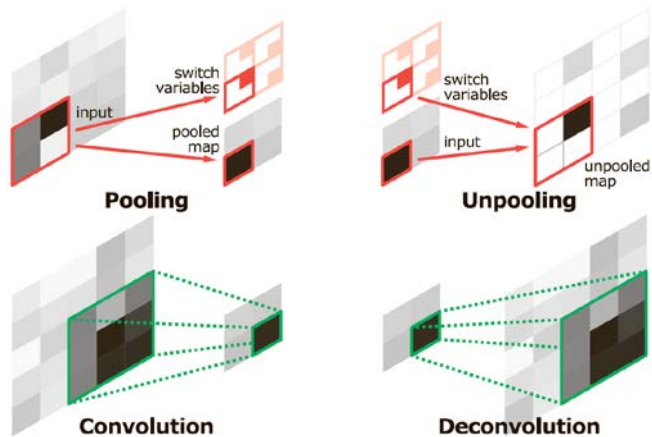

**Fig. 1. Illustration of deconvolution and unpooling layers [43]**

We have used the following practical rules of thumb in creating our CAE models in Caffe:

1.  The main issue with a deep AE is asymmetry [27]. Thus, the model should be symmetric in terms of the total size of the feature maps and the number of neurons in all hidden layers in both the encoder and decoder parts. These sizes and numbers should decrease from layer to layer in the encoder part and increase in the same way in the decoder part providing the encoder-decoder paradigm, i.e. contractive AE [19]. These sizes and numbers should not be less than some minimal values, allowing handling the size of the input problem from the informational point of view;

2.  As a cost function, it is better to use two loss layers, <*Sigmoid_Cross_Entropy_Loss*> and <*Euclidean_Loss*> [5, 19]. Preliminary experiments has shown that the use of only one of these loss layers separately does not provide good training convergence:

- the cross-entropy (logistic) loss layer is defined by the expression

$$E = -\frac{1}{N} \sum_{n=1}^{N} \left[ y_n \log \hat{y}_n + (1 - y_n) \log(1 - \hat{y}_n) \right], \qquad (1)$$

where $N$ is a number of samples, $y$ are the targets $y \in \{0,1\}$, $\hat{y}_n \equiv f(w \cdot x_n)$, where $f$ is a logistic function, $w$ is the vector of weights optimized through gradient descent and $x$ is the input vector [44]. and,

- the Euclidean (L2) loss layer defined by expression

$$E = \frac{1}{2N} \sum_{n=1}^{N} \left\| \hat{y}_n - y_n \right\|_2^2,$$

where $\hat{y}$ are the predictions $\hat{y} \in \{-\infty, +\infty\}$ and $y$ are the targets $y \in \{-\infty, +\infty\}$.

3. Visualization of the values (along with numerical representations) of trainable filters, feature maps and hidden units from layer to layer plays a very important diagnostic role. It allows us to inspect the function of intermediate layers and, therefore, better understand how data are converted/processed inside a deep model [45];

4. The main purpose of an activation function after each layer is non-linear data processing [15]. Since the nature of convolution/deconvolution operations is a multiplication, our visualization showed that the result of the convolution/deconvolution operations (the values of feature maps) increasing sharply from layer to layer, preventing the CAE model from converging during training. Thus, the use of sigmoid or hyperbolic tangent activation functions, which constrain the resulting values of feature maps to the interval [0,1] or [-1,1] respectively, sets appropriate limits on the values of feature maps at the end of the decoder part, and provides good convergence of the whole model;

5. The well-known fact is that good generalization properties of NNs depend on the ratio of trainable parameters, i.e. weights (|$w$|) and biases (|$b$|) to the size of the input data (|$Data$|). Therefore experimentation is required [5, 46] to find the AE architecture, (i.e. the size of trainable parameters and appropriate number of neurons in all hidden layers) with the best generalization properties. Then, a similar size in terms of the total size of feature maps and the number of neurons in all hidden layers could be considered as an appropriate starting point when designing a CAE with good generalization properties. Direct comparison of AE and CAE models is inappropriate, because a CNN of the same size as a given fully-connected network would have fewer trainable parameters [6];

6. The created model should be stable. NN training may converge to different local optima in different runs depending on initial weights/biases. By "stable model" we mean that the same convergence results can be obtained in several consecutive runs (at least three).

We have created five deep CAE models in Caffe, denoted in Table 1, as follows:

Model 1 (Fig. 2 (left)), notation (*conv <-> deconv*), contains two convolutional layers followed by two fully-connected layers in the encoder part and, inversely, one fully-connected layer followed by two deconvolution layers in the decoder part. This is the model which we have investigated in our arXiv paper [14];

Model 2 (Fig. 2 (right)), notation (*conv, pool <-> deconv*), contains two pairs of convolutional and pooling layers followed by two fully-connected layers in the encoder part and, inversely, one fully-connected layer followed by only two deconvolution layers in the decoder part. As we can see from Table 1, Model 2 is not symmetric. However, we have included it because we have seen a similar idea in the Neon deep learning framework [41] and wanted to study it;

Model 3 (Fig. 3 (left)), notation (*conv, pool, sw (switches) <-> deconv, sw, unpool*), contains two pairs of convolutional and pooling layers followed by two fully-connected layers in the encoder part and, inversely, one fully-connected layer followed by two pairs of deconvolution and unpooling layers WITH the use of switch variables in the decoder part. The unpooling layer works here as follows: the max-pooled feature (what) is restored into the correct place (where) within the unpooled output feature map, as specified by the switch variable. All other elements of the map are zeros. Model 3 is similar to the networks presented in [27, 29];

Model 4 (Fig. 3 (right)), notation (*conv, pool <-> deconv, unpool*), contains two pairs of convolutional and pooling layers followed by two fully-connected layers in the encoder part and, inversely, one fully-connected layer followed by two deconvolution and unpooling layers WITHOUT switch variables in the decoder part. The unpooling layer works here as follows: the max-pooled feature (what) is restored into a predetermined place (where) within the unpooled output feature map, all other elements of the map are zeros. This approach could be called "centralized" unpooling, when the max-pooled feature is placed in the center of each pool [29, 47]. In the Noh implementation [42] we used, this specific position is a left and a top corner [0,0] of each pool. Some research results for this unpooling mode are presented in [29];

Model 5 (does not have a Figure), notation (*conv, pool (tanh) <-> deconv, unpool (tanh)*), the same model as the Model 4 except of using a hyperbolic tangent activation function in all layers. All previous Model 1 – Model 4 use a standard sigmoid activation function.

More technical details about our Models (with the reference to Table 1 and Figs. 2-3) along with the results of their experimental research are presented in the next Section.

**Fig. 2. Graphical representation of Model 1 (left) and Model 2 (right)**

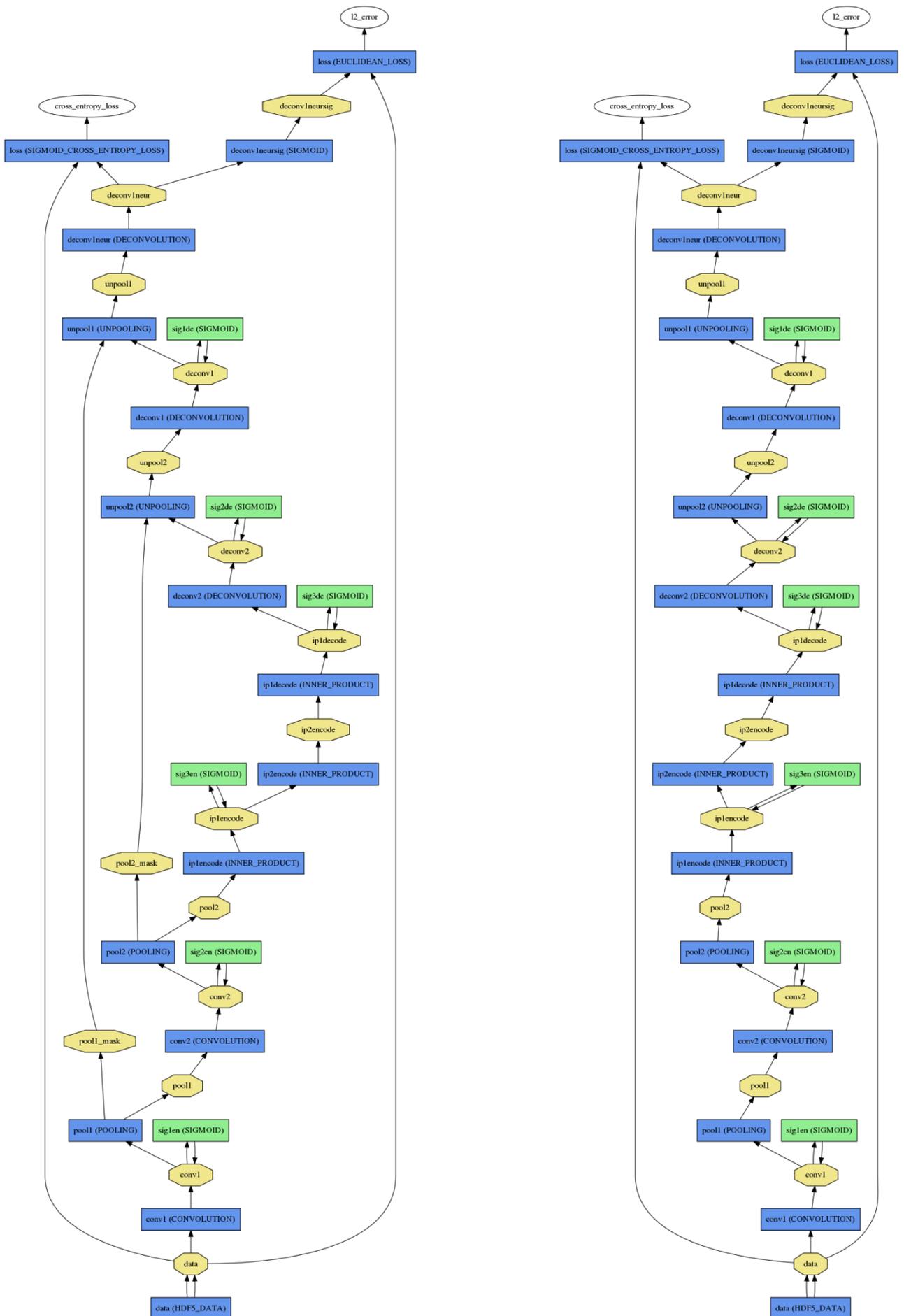

Fig. 3. Graphical representation of Model 3 (left) and Model 4 (right)

# 4. Experimental results

## 4.1. Size of the models

We have used the MNIST dataset [48] for the experimental research. In all experiments we have used the standard 60000 examples as a training set and 10000 examples as a testing set. The size of the MNIST training dataset is 60000 examples x 784 elements, so our $|Data|$ = 47040 K elements. We will be using this number to calculate the ratio of trainable parameters of our Models (weights $|w|$ and biases $|b|$) to the size of the input data $|Data/(w+b)|$ in our experiments below. We have created an HDF5 version of the MNIST dataset and have not applied any modification to input images except for normalization into the range [0,1]. We have used the MNIST labels only for visualization of clustering results.

In the Caffe examples there are two models which solve a dimensionality reduction task: a semi-supervised Siamese network [49], proposed by Hadsell et al. [50] and a deep AE [46], proposed by Hinton et al. [5]. We have included these two models in our results for comparison with our Models 1-5. A Siamese network consists of two coupled LeNet [6] architectures followed by a contrastive loss function. It trains in a semi-supervised fashion, since we form the training set by labelling a pair of input images (chosen randomly) as 1 if the images belong to the same class and 0 otherwise. We have placed the Siamese model in the first row of Table 1 to have a semi-supervised model first in the list and followed by all unsupervised models. In the Siamese network available in the examples [49] we have changed the number of planes (feature maps) in the convolutional layers and the number of neurons in the fully-connected layers to be the same as the encoder part of our Models 2-5. We left only one ReLU (a Rectified Linear Unit activation function) layer before the fully connected layers as in the Caffe examples [49].

According to our rule 5 in Section 3 above, we have researched a deep AE [5] in our previous paper [14]. We have shown that the exact deep AE architecture (1000-500-250-30) presented in [5] provides the best generalization properties out of 5 different models. This model is specified in the second row of Table 1. Rule 5 says that a CAE model with the same total size of feature maps and the same number of neurons in all hidden layers (we call this parameter "Model size", see the fourth column, Table 1) could be considered as an appropriate starting point when designing the model with good generalization properties. However, the research results of CAE models without pooling – unpooling layers in [14] have actually shown that a CAE model which is twice as big (in terms of feature maps) gives better results. This CAE model is our Model 1 and it is specified in the third row of Table 1. Model 1 has about 297K training parameters and its $|Data/(w+b)|$ ratio is 158. Therefore we have created our Models 2-5 trying to keep the same $|Data/(w+b)|$ ratio which is 148 and 139 for the Model 2 and the Model 3 respectively. Note that our Models 1-5 are half the size (in terms of trainable parameters) of the SWWAE with 679K trainable parameters [27].

Table 1 contains the architecture parameters for all researched models. Table 2 contains a detailed calculation of the number of trainable parameters for a Siamese network. Table 3 contains a detailed calculation of the number of trainable parameters for a deep AE and Models 1-3 only, because Models 4-5 parametrically are similar to Model 3. We trained each model in three fashions, with 2, 10 and 30 (30 is exactly as in Caffe examples) neurons in the last hidden layer of the encoder part $<ip2encode>$, which corresponds to 2-, 10-, 30-dimensional (we will be using abbreviations 2D, 10D and 30D) space of encoding respectively. Here in Tables 1-3 all calculations are provided for the 2D case. But graphically the layer $<ip2encode>$ looks a bit different (Figs. 2-3): since it does a linear inner product operation only, it does not have a green rectangle with activation function; find a blue rectangle $<ip2encode>$ somewhere in the center of each Model. The followed blob with the same title, see above the yellow octagon $<ip2encode>$, keeps the result of this operation. This is an actual place, i.e. 2, 10 and 30 neurons, where data are stored. Further we will be calling it as "internal code" of our Models. Also it serves as an input for the decoder part. Therefore, graphically we see that the Models have two fully-connected layers $<ip1encode>$ and $<ip2encode>$ in the encoder part and only one fully-connected layer $<ip1decode>$ in the decoder part. We, therefore, use this rhetoric, that our Models have two fully-connected layers in the encoder part and only one fully-connected layer in the decoder part. However, it provides a symmetric number of feature maps and the number of neurons in the hidden layers (see the third column of Table 1, bold).

To provide a better understanding of our notations in Table 1 as well as Figs. 2-3, let us explain the $conv/deconv$ layers for Model 1 and Model 3. The encoder part of Model 1 has 8 planes (or feature maps) with kernel 9x9 in the $conv1$ layer and 4 planes with kernel 9x9 in the $conv2$ layer. The decoder part has 4 planes with kernel 12x12 in the $deconv2$ layer and 4 planes with kernel 17x17 in the $deconv1$ layer. Model 3 has 26 planes with kernel 5x5 in the $conv1$ layer followed by the max-pooling layer $pool1$ with kernel 2x2 and 36 planes with kernel 5x5 in the $conv2$ layer followed by the max-pooling layer $pool2$ with kernel 2x2 in the encoder part. In the decoder part Model 3 has 36 planes with kernel 4x4 in the $deconv2$ layer followed by the unpooling layer $unpool2$ with kernel 2x2 and 26 planes with kernel 5x5 in the $deconv1$ layer followed by the unpooling layer $unpool1$ with kernel 2x2. In the decoder part, the purpose of the deconvolution layer $deconv1neur$, which corresponds to the term $(1*1w+1b)$ for Model 1 and the term $(5*5w+1b)$ for Model 3 in the third column of Table 3, is to reshape all feature maps of the last decoder layer $deconv1$ or $unpool1$ into one reconstructed image with the same size of 28x28 pixels as the original MNIST image. There was an explanation in the Caffe user group how to do that [51].

Since the deconvolution operation has the same nature as convolution [29], we have used the same approach to calculate the number of trainable parameters both in the encoder and decoder parts (Table 3). The accuracy of our calculations can be easily checked by calling Caffe from Matlab or Python using the command $<caffe('weights');>$. As we can see from Tables 1 and 3, our Models (except of the Model 2) are practically symmetric not only in terms of the total number of elements in feature maps and the number of neurons in the hidden layers, but also in terms of the number of trainable parameters in both the encoder and decoder parts.

**Table 1. Architecture parameters of researched models**

| Model | Architecture | Size of feature maps and number of hidden nodes | Model size, elements | Number of trainable parameters $|w+b|$ |
|---|---|---|---|---|
| Siamese network | Each channel: 784-*conv1*(5x5x26)-*pool1*(Max2x2)-*conv2*(5x5x36)-*pool2*(Max2x2)-250-2 | *conv1*(24x24x26)-*pool1*(12x12x26)-*conv2*(8x8x36)-*pool2*(4x4x36)-250-2 <br><br> **(14976-3744-2304-576-250-2) * 2** | 43704 | 338K |
| Deep AE | 784-1000-500-250-2-250-500-1000-784 | 1000-500-250-2-250-500-1000 <br> **1000-500-250-2-250-500-1000** | 3502 | 2823K |
| <u>Model 1</u>, CAE <br> *conv <-> deconv* | 784-*conv1*(9x9x8)-*conv2*(9x9x4)-250-2-250-*deconv2*(12x12x4)-*deconv1*(17x17x4)-784 | *conv1*(20x20x8)-*conv2*(12x12x4)-250-2-250-*deconv2*(12x12x4)-*deconv1*(28x28x4) <br> **3200-576-250-2-250-576-3136** | 7990 | 297K |
| <u>Model 2</u>, CAE <br> *conv, pool <-> deconv* | 784-*conv1*(5x5x26)-*pool1*(Max2x2)-*conv2*(5x5x36)-*pool2*(Max2x2)-250-2-250-*deconv2*(12x12x4)-*deconv1*(17x17x4)-784 | *conv1*(24x24x26)-*pool1*(12x12x26)-*conv2*(8x8x36)-*pool2*(4x4x36)-250-2-250-*deconv2*(12x12x4)-*deconv1*(28x28x4) <br> **14976-3744-2304-576-250-2-250-576-3136** | 25814 | 318K |
| <u>Model 3</u>, CAE <br> *conv, pool, sw (switches) <-> deconv, sw, unpool* | 784-*conv1*(5x5x26)-*pool1*(Max2x2)-*conv2*(5x5x36)-*pool2*(Max2x2)-250-2-250-*deconv2*(4x4x36)-*unpool2*(Max2x2)-*deconv1*(5x5x26)-*unpool1*(Max2x2)-784 | *conv1*(24x24x26)-*pool1*(12x12x26)-*conv2*(8x8x36)-*pool2*(4x4x36)-250-2-250-*deconv2*(4x4x36)-*unpool2*(8x8x36)-*deconv1*(12x12x26)-*unpool1*(24x24x26) <br> **14976-3744-2304-576-250-2-250-576-2304-3744-14976** | 43702 | 338K |
| <u>Model 4</u>, CAE <br> *conv, pool <-> deconv, unpool* | Same as <u>Model 3</u> | Same as <u>Model 3</u> | 43702 | 338K |
| <u>Model 5</u>, CAE <br> *conv, pool (tanh) <-> deconv, unpool (tanh)* | Same as <u>Model 3</u> | Same as <u>Model 3</u> | 43702 | 338K |

**Table 2. Calculation of the number of trainable parameters for Siamese network**

| Model | Number of trainable parameters, w(weights)+b(biases), i(inputs), o(outputs) | | |
|---|---|---|---|
| | Left channel | Right channel | Total |
| Siamese network | *conv1*->((5*5w+1b)*1i*26o)+ <br> *conv2*->((5*5w+1b)*26i*36o)+ <br> *ip1encode*->(576i*250o+250b)+ <br> *ip2encode*->(250i*2o+2b) = <br> (650w+26b)+(23400w+36b)+ <br> (144000w+250b)+(500w+2b) = ***168864*** | *conv1*->((5*5w+1b)*1i*26o)+ <br> *conv2*->((5*5w+1b)*26i*36o)+ <br> *ip1encode*->(576i*250o+250b)+ <br> *ip2encode*->(250i*2o+2b) = <br> (650w+26b)+(23400w+36b)+ <br> (144000w+250b)+(500w+2b) = ***168864*** | ***337728*** <br> ***(338K)*** |

**Table 3. Calculation of the number of trainable parameters for unsupervised models**

| Model | Number of trainable parameters, w(weights)+b(biases), i(inputs), o(outputs) | | |
|---|---|---|---|
| | Encoder part | Decoder part | Total |
| Deep AE | *encode1*->(784i\*10000+1000b)+ *encode2*->(1000i\*500o+500b)+ *encode3*->(500i\*250o+250b)+ *encode4*->(250i\*2o+2b)= (784000w+1000b)+(500000w+500b)+ (125000w+250b)+(500w+2b) = **1411252** | *decode4*->(2i\*250o+250b)+ *decode3*->(250i\*500o+500b) + *decode2*->(500i\*1000o+1000b)+ *decode1*->(1000i\*784o+784b)= (500w+250b)+(125000w+500b)+ (500000w+1000b)+(784000w+784b) = **1412034** | ***2823286*** ***(2823K)*** |
| Model 1 | *conv1*->((9\*9w+1b)\*1i\*8o)+ *conv2*->((9\*9w+1b)\*8i\*4o)+ *ip1encode*->(576i\*250o+250b)+ *ip2encode*->(250i\*2o+2b) = (648w+8b)+(2592w+4b)+ (144000w+250b)+(500w+2b) = **148004** | *ip1decode*->(2i\*250o+250b)+ *deconv2*->((12\*12w+1b)\*250i\*4o)+ *deconv1*->((17\*17w+1b)\*4i\*4o)+ *deconv1neur*->((1\*1w+1b)\*4i\*1o)+ (500w+250b)+(144000w+4b)+ (4624w+4b)+(4w+1b) = **149387** | ***297391*** ***(297K)*** |
| Model 2 | *conv1*->((5\*5w+1b)\*1i\*26o)+ *conv2*->((5\*5w+1b)\*26i\*36o)+ *ip1encode*->(576i\*250o+250b)+ *ip2encode*->(250i\*2o+2b) = (650w+26b)+(23400w+36b)+ (144000w+250b)+(500w+2b) = **168864** | *ip1decode*->(2i\*250o+250b)+ *deconv2*->((12\*12w+1b)\*250i\*4o)+ *deconv1*->((17\*17w+1b)\*4i\*4o)+ *deconv1neur*->((1\*1w+1b)\*4i\*1o)+ (500w+250b)+(144000w+4b)+ (4624w+4b)+(4w+1b) = **149387** | ***318251*** ***(318K)*** |
| Model 3 | *conv1*->((5\*5w+1b)\*1i\*26o)+ *conv2*->((5\*5w+1b)\*26i\*36o)+ *ip1encode*->(576i\*250o+250b)+ *ip2encode*->(250i\*2o+2b) = (650w+26b)+(23400w+36b)+ (144000w+250b)+(500w+2b) = **168864** | *ip1decode*->(2i\*250o+250b)+ *deconv2*->((4\*4w+1b)\*250i\*36o)+ *deconv1*->((5\*5w+1b)\*36i\*26o)+ *deconv1neur*->((5\*5w+1b)\*26i\*1o)+ (500w+250b)+(144000w+36b)+ (23400w+26b)+(650w+1b) = **168863** | ***337727*** ***(338K)*** |

### *4.2. Dimensionality reduction results*

The experimental results of a Siamese network and a deep AE along with Models 1-5 are presented in Table 4. We did three runs for each model. For unsupervised models the values of both loss layers <*Sigmoid_Cross_Entropy_Loss*> and <*Euclidean_Loss*> are separated by semicolon. We presented a value of a <*Contrastive_Loss*> layer for a Siamese network. The values of loss layers are presented in red for the training set and in black for the test set. The learning parameters of the solver for all models were the same: 50000 training iterations were used, Stochastic Gradient Descent algorithm used one hundred patterns in a batch, base learning rate was 0.006, learning policy was "fixed", and weight decay was equal to 0.0005. A standard sigmoid activation function was used in the deep AE (it was in examples) and our Models 1-4. A standard hyperbolic tangent activation function was used in Model 5. We ran many experiments changing the architecture and learning parameters, but in many cases they were not stable architectures. For example, we tried different initializations of weights and biases (<*weight_filler*> and <*bias_filler*>). The presented results were obtained with the following initialization: <*bias_filler {type: "constant"}*> for all layers, <*weight_filler {type: "xavier"}*> for convolutional/deconvolution layers, <*weight_filler {type: "gaussian" std: 1 sparse: 25}*> for fully-connected (InnerProduct) layers.

In order to estimate the quality of feature extraction in general, and the quality of the encoded 2D, 10D and 3D internal code in particular, we have used this internal code as an input to a classifier as in literature [23-29]. The goal of this experiment was to apply the classifiers with the same structure and learning parameters and estimate which Model provides the best internal code as determined by classification results. We have used two simple classification models, linear model (function <*classify()*>) and a multi-layer perceptron (MLP) implemented in the Matlab Statistics and Machine Learning Toolbox and Neural Network Toolbox respectively. We created three separate MLP models for 2D, 10D and 30D internal code keeping the ratio $|Data/(w+b)| = 100$ (where $|Data|$ is, for example, 20K for the 2D internal code, given the MNIST test set contains the 10000 instances). Thus we have used three models, 2-14-10, 10-42-10 and 30-66-10 for these three cases. Standard Matlab functions were used: <*patternnet()*> - to create a MLP, <*train()*> - to train it and <*net()*> - to provide classification results. We used sigmoid and softmax activation functions for hidden and output layers of the MLP respectively in all cases. The following learning parameters were used: the training accuracy was $10^{-10}$, the minimum training gradient was $10^{-10}$, training epochs were 6000 and the maximum training validations checks were 6000. A scaled conjugate gradient backpropagation algorithm '*trainscg*' was used for the training. We used 10-fold cross-validation and calculated an average per-class classification error [52]. The classification results are specified in the seventh and eighth columns of Table 4. The last column of Table 4 contains the sparsity estimation of the internal code for all researched models (see more details in Section 5.1).

Obviously, a MLP has shown better classification results than a linear model. A semi-supervised Siamese network showed the best classification results (1.39% by 10D Siamese network) in comparison with unsupervised models. Model 1 (**2.65%** by 30D Model 1) and Model 2 (**2.54%** by 30D Model 2) have provided slightly better results than a deep AE (2.85% by 30D deep AE). Model 3 with pooling-unpooling layers and WITH switches obtained the worst classification results (8.75% and 3.75% by 10D and 30D Model 3). Model 4, with pooling-unpooling layers and WITHOUT switches, has shown the best classification results (**2.19%** by 30D Model 4). The results of Model 5 (3.36% by 10D Model 5) show that the use of a hyperbolic tangent activation function provides worse results than the use of a sigmoid activation function. For comparison we also run the same classification experiment using an MLP with the same number of trainable parameters, the same learning parameters and the same 10-fold cross validation on original 784-dimensional MNIST images. The MLP 784-88-10 (which has same ratio $|Data/(w+b)| = 100$) obtained worse **2.85%** average per-class classification error. Therefore we can conclude, that the developed Models provide very good quality feature extraction and dimensionality reduction on the MNIST dataset, because the obtained classification errors are less than 12% in a case of a linear model and less than 2.85% in a case of a MLP.

**Table 4. Performances of the training, average per-class classification errors and sparsity of internal code**

| Model, Number of trainable parameters $\|w+b\|$, Model size (elements) | Data/ (w+b) | Values of loss layers, (*Sigmoid_Cross_Entropy_Loss*; *Euclidean_Loss*), (*Contrastive_Loss*|Siamese ), train / test | | | | Average per-class classification error, % | | Sparsity of internal code, % |
|---|---|---|---|---|---|---|---|---|
| | | Dims | Run 01 | Run 02 | Run 03 | Linear | MLP | |
| Siamese network, 338K, 43704 | 139 | 2 | 0.031 / 0.026 | 0.019 / 0.016 | 0.021 / 0.017 | 3.49 | 3.35 | 59.3 |
| | | 10 | 0.028 / 0.024 | 0.018 / 0.016 | 0.017 / 0.016 | **1.48** | **1.39** | 69.1 |
| | | 20 | 0.036 / 0.026 | 0.022 / 0.015 | 0.020 / 0.016 | 1.62 | 1.48 | 79.6 |
| Deep AE, 2823K, 3502 | 17 | 2 | 132.07; 14.20 138.92; 15.07 | 129.14; 13.67 137.11; 14.78 | 132.75; 14.31 140.56; 15.40 | 22.68 | 13.52 | 77.6 |
| | | 10 | 84.07; 5.91 86.64; 6.19 | 83.18; 5.79 86.71; 6.21 | 82.21; 5.59 86.34; 6.15 | **6.91** | 3.46 | 70.3 |
| | | 30 | 66.56; 3.11 68.72; 3.31 | 66.23; 3.04 68.51; 3.27 | 66.48; 3.10 68.50; 3.27 | 7.46 | **2.85** | 78.9 |
| Model 1, 297K, 7990 | 158 | 2 | 137.79; 15.23 143.59; 15.93 | 137.69; 15.26 144.72; 16.09 | 136.15; 14.88 143.52; 15.87 | 19.82 | 13.73 | 78.3 |
| | | 10 | 83.11; 5.71 86.19; 6.05 | 85.96; 6.29 88.35; 6.48 | 85.06; 6.05 86.21; 6.05 | **6.30** | 3.48 | 69.7 |
| | | 30 | 68.47; 3.47 69.34; 3.44 | 69.04; 3.49 69.65; 3.44 | 71.43; 3.88 73.72; 4.08 | 6.88 | **2.65** | 79.7 |
| Model 2, 318K, 25814 | 148 | 2 | 137.58; 15.28 146.27; 16.49 | 138.73; 15.36 144.23; 16.03 | 137.51; 15.08 144.80; 16.15 | 21.33 | 14.38 | 79.7 |
| | | 10 | 86.93; 6.26 88.13; 6.46 | 85.12; 6.03 86.12; 5.99 | 83.58; 5.78 85.36; 5.92 | **5.31** | 3.77 | 75.8 |
| | | 30 | 64.64; 2.83 66.02; 2.90 | 65.28; 2.93 66.38; 2.96 | 69.58; 3.58 70.88; 3.69 | 6.41 | **2.54** | 85.0 |
| Model 3, 338K, 43702 | 139 | 2 | 54.18; 1.13 55.75; 1.15 | 55.31; 1.18 55.30; 1.08 | 55.38; 1.21 56.36; 1.22 | 45.72 | 41.53 | 30.8 |
| | | 10 | 51.67; 0.75 52.45; 0.74 | 51.62; 0.74 52.49; 0.76 | 51.81; 0.78 52.48; 0.76 | 13.75 | 8.75 | 67.4 |
| | | 30 | 51.21; 0.70 51.97; 0.69 | 51.15; 0.69 51.87; 0.68 | 51.34; 0.71 52.19; 0.72 | **6.82** | **3.75** | 74.8 |
| Model 4, 338K, 43702 | 139 | 2 | 135.10; 14.43 142.15; 15.43 | 136.56;14.77 142.56; 15.52 | 135.08; 14.35 141.99; 15.39 | 21.23 | 13.87 | 70.9 |
| | | 10 | 81.38; 5.29 83.23; 5.42 | 84.07; 5.67 84.82; 5.68 | 82.78; 5.48 84.33; 5.60 | **5.06** | 2.89 | 73.7 |
| | | 30 | 64.19; 2.45 65.35; 2.55 | 65.94; 2.76 66.83; 2.77 | 64.03; 2.48 66.10; 2.72 | 5.76 | **2.19** | 86.9 |
| Model 5, 338K, 43702 | 139 | 2 | 141.72; 15.60 150.56; 16.81 | 138.05; 14.96 141.70; 15.38 | 139.70; 15.27 146.24; 16.15 | 20.63 | 12.34 | 69.7 |
| | | 10 | 86.08; 6.02 90.66; 6.62 | 86.35; 6.10 91.05; 6.68 | 86.89; 6.19 91.00; 6.71 | **6.14** | **3.36** | 68.9 |
| | | 30 | 75.62; 4.31 78.75; 4.71 | 78.64; 4.61 72.18; 3.76 | 75.08; 4.22 77.40; 4.51 | 6.61 | 3.44 | 77.6 |

We also collected the training times for unsupervised models only (for 50000 training iterations) both in CPU and GPU modes (Table 5) on the three computational systems located in the Canadian Centre for Behavioural Neuroscience (CCBN), University of Lethbridge, Canada:

- The workstation *WS1241* operated under Ubuntu 14.04.2. It is equipped with a 4-core (total of 8 CPU threads visible in Linux) Inter(R) Xeon(R) E5620@2.40 GHz processor, 51 Gb of RAM and an NVidia(R) GeForce GTS 450 GPU. The GPU has Fermi architecture, 192 CUDA cores, 1 Gb of RAM and compute capability 2.1;
- The workstation *Polaris1* operated under Ubuntu 16.04.2. It is equipped with two 8-core (total of 32 CPU threads visible in Linux) Inter(R) Xeon(R) E5-2630 v3 @2.40 GHz processors, 256 Gb of RAM and an NVidia(R) GeForce GTX TITAN X GPU. The GPU has Maxwell architecture, 3072 CUDA cores, 12 Gb of RAM and compute capability 5.2;
- The cluster *Hodgkin* operated under Red Hat Enterprise Linux. The cluster consists of 64 blade servers. Each server is equipped with two 6-core Intel(R) Xeon(R) CPU E5520@2.27 GHz processors and 48 GB of RAM. The cluster has a total of 768 CPU cores and 3072 GB of RAM. We have used *Hodgkin* in a CPU mode only. Since it has a lot of CPU cores we run many models (3 times per model) in parallel and, therefore, we have received most of our results on this computational platform.

Interestingly, as we can see from Table 5, the relation between training times is straightforward in CPU computational mode, where the bigger Models 3-5 obviously train longer. But the training times in GPU mode are smaller (practically 1.5 times) for the bigger Models 3-5, which have larger feature maps in *conv/deconv* layers, and more trainable parameters (see Tables 1-2) in comparison with Models 1-2. This is because a GPU-implementation of *conv/deconv* layers in Caffe is much better optimized (i.e. bigger-sized layers are computed much faster) than appropriate CPU implementation thanks to NVidia CUDA technology [53] and the cuDNN library [54].

**Table 5. Average training times of one run of unsupervised models (minutes)**

| System | Deep AE | Model 1 | Model 2 | Model 3 | Model 4 | Model 5 |
|---|---|---|---|---|---|---|
| *Hodgkin* CPU Intel(R) Xeon(R) E5520@2.27GHz | 129 | 478 | 702 | 755 | 758 | 762 |
| *WS1241* GPU GeForce GTS 450 | 18 | 249 | 239 | 141 | 142 | 140 |
| *Polaris1* GPU GeForce GTX TITAN X | 6 | 91 | 83 | 57 | 60 | 58 |

*4.3. Clustering and visualization results*

We have used the t-SNE technique [55] to visualize 10D and 30D internal code, produced by all models. The t-SNE technique is an advanced method for dimensionality reduction and visualization. It is based on Stochastic Neighbor Embedding (SNE) [56] which converts the high-dimensional Euclidean distances between datapoints into conditional probabilities that represent similarities. SNE minimizes the sum of Kullback-Leibler divergences over all datapoints using a gradient descent method. t-SNE differs from SNE in two ways: (1) it uses a symmetrized version of the SNE cost function with simpler gradients and (2) it uses a Student-t distribution rather than a Gaussian to compute the similarity between two points in the low-dimensional space. This allows t-SNE to have a cost function that is easier to optimize and produces significantly better visualizations [55]. This advanced technology has been used for visualization in many state-of-the-art problems, for example, to visualize the last hidden-layer representations of a deep Q-network playing Atari 2600 games [57].

Our clustering task is to encode the 10,000 instances of the MNIST test set in a 2D space. A Siamese network does it in a semi-supervised fashion, and a deep AE and our Models 1-5 do it in an unsupervised fashion. The visualization results from our previous paper [14] have shown that there is no big difference between visualized 10D and 30D internal codes. Therefore we are presenting here the 10D visualizations only. The visualizations of how a Siamese network, a deep AE, and our Models 1-5 solve the clustering task are depicted in Figs. 4-10 respectively. Table 6 shows how a deep AE and our Models 1-5 provide the reconstruction of several example MNIST images.

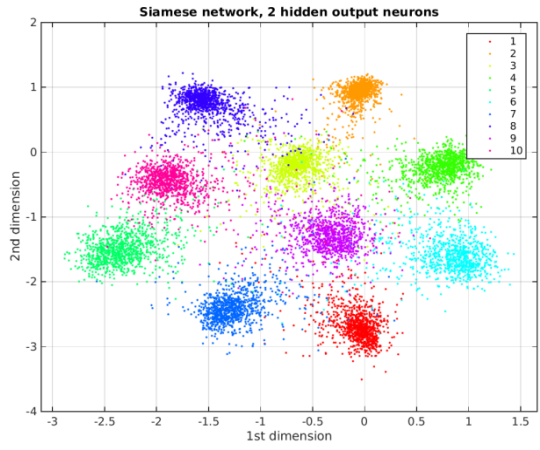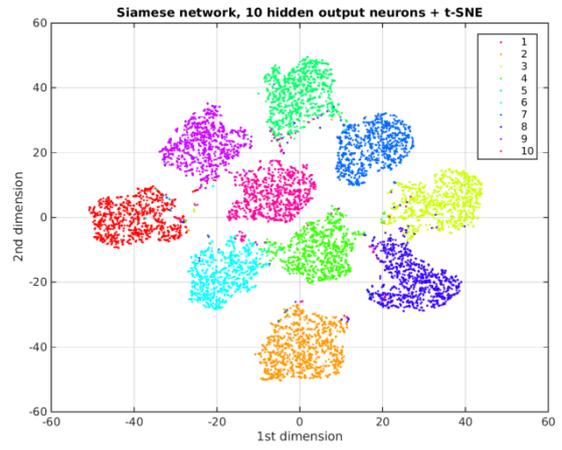

**Fig. 4. Visualization of MNIST test set in a 2D space by 2D (left) and 10D (right) Siamese network**

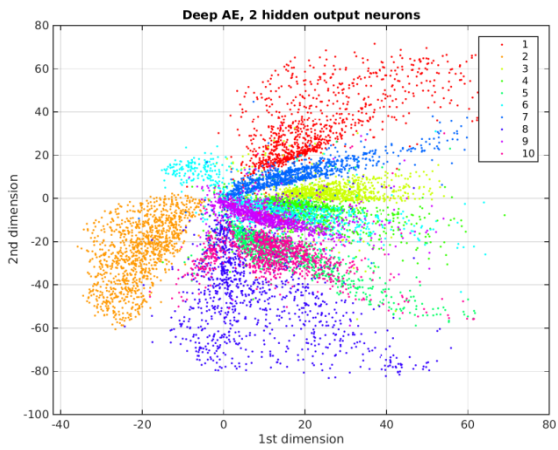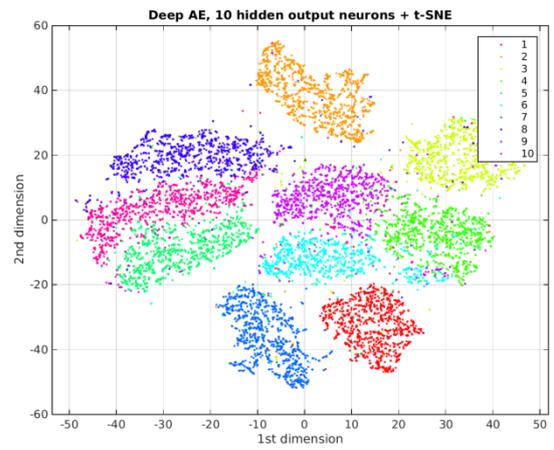

**Fig. 5. Visualization of MNIST test set in a 2D space by 2D (left) and 10D (right) deep AE**

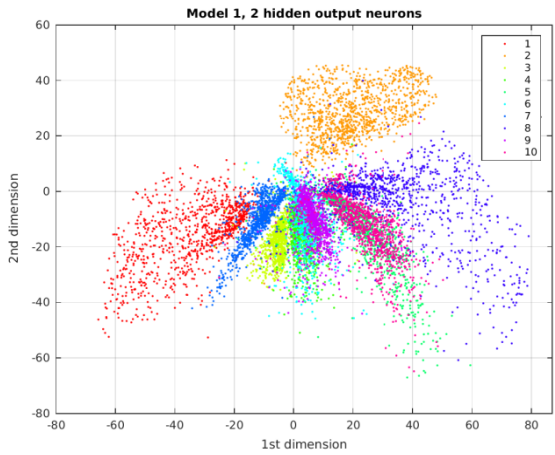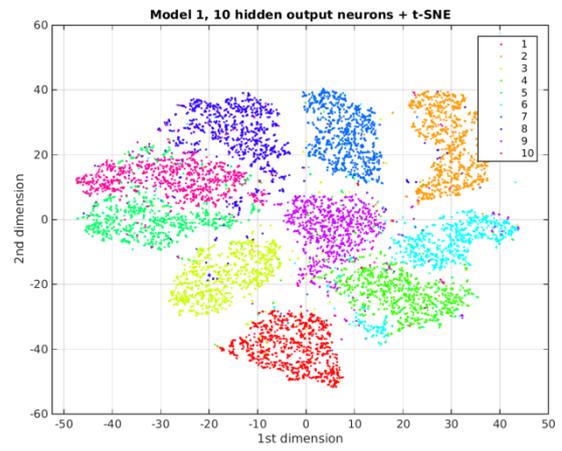

**Fig. 6. Visualization of MNIST test set in a 2D space by 2D (left) and 10D (right) Model 1**

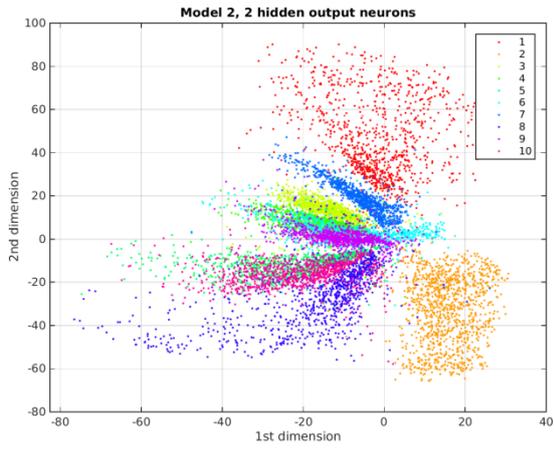 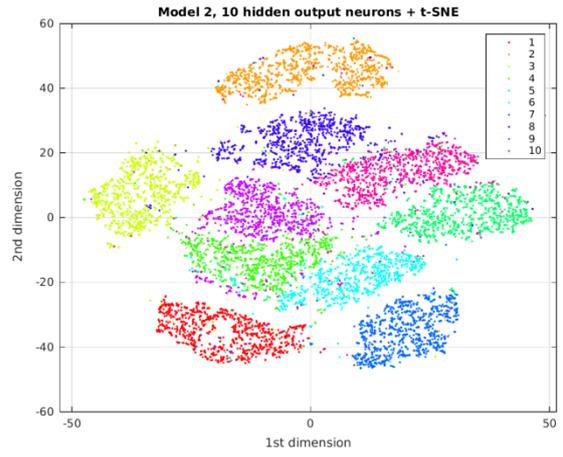

**Fig. 7. Visualization of MNIST test set in a 2D space by 2D (left) and 10D (right) Model 2**

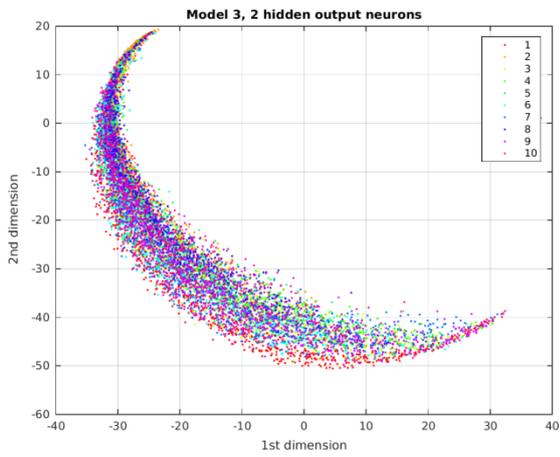 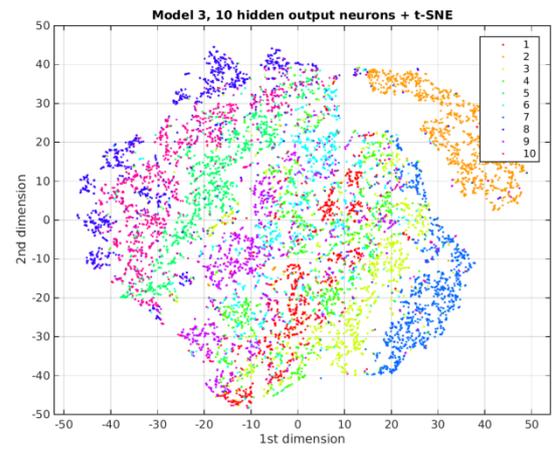

**Fig. 8. Visualization of MNIST test set in a 2D space by 2D (left) and 10D (right) Model 3**

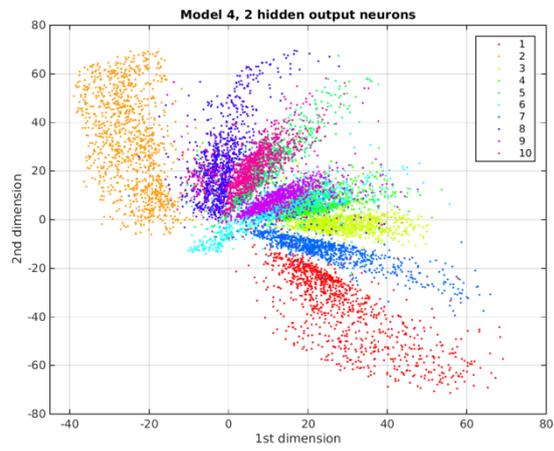 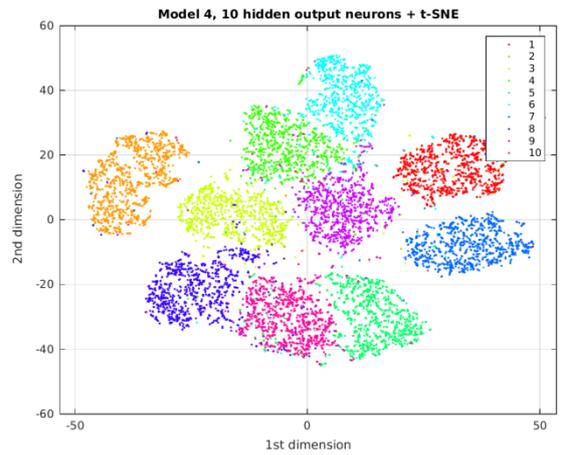

**Fig. 9. Visualization of MNIST test set in a 2D space by 2D (left) and 10D (right) Model 4**

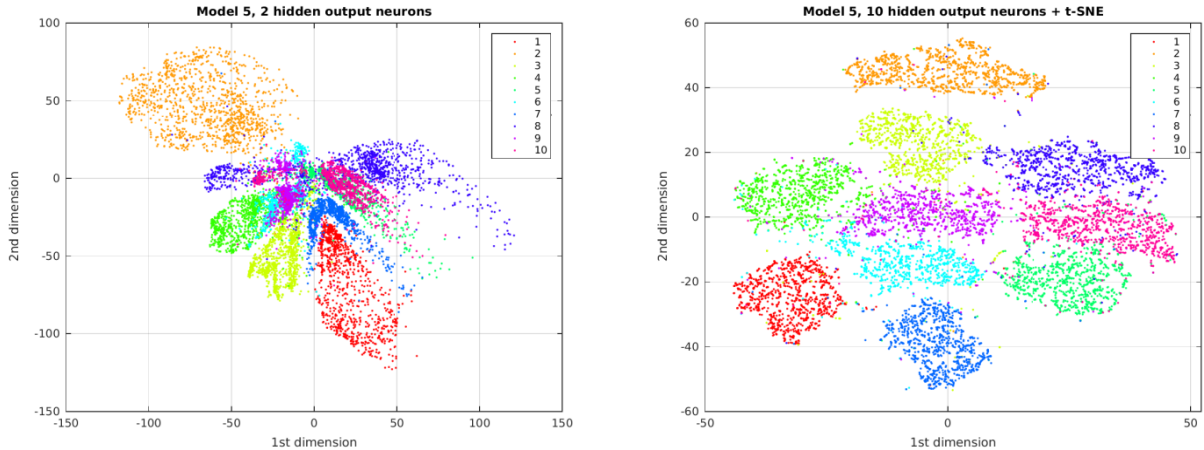

**Fig. 10. Visualization of MNIST test set in a 2D space by 2D (left) and 10D (right) Model 5**

The clustering results in Figs. 4-10 and reconstruction results in Table 6 confirm the numerical results (which characterize the quality of the internal codes) in Table 1. Model 4, which provided the smallest classification errors, has (not surprisingly) shown the best clustering results (Fig. 9) among the unsupervised models. The clusters produced by Model 4 are more dense (Fig. 9 (right)) and look qualitatively similar to the result of the semi-supervised Siamese model (Fig. 4 (right)) comparing to the other models. Model 4 has also shown the second-best reconstruction quality of MNIST images among the unsupervised models, after Model 3 (Table 6). Model 3 has shown contradictory results. It provided the worst classification errors in Table 1 and the worst (even unacceptable) clustering results (Fig. 8). However, it has shown the smallest values of our cost function (Table 1) and practically ideal reconstruction quality of MNIST images (Table 6). We discuss these results further in Section 5.1.

As mentioned above in rule 3, the visualization of a deep network plays very important diagnostic role and allows us to easily inspect the function of intermediate layers. Similarly to Zeiler and Fergus [45], who stated that visualization allowed them to find model architectures that outperform Krizhevsky et al. on the ImageNet classification benchmark, our visualization allowed us basically to create our Models. The implementation of such visualization is simple and straightforward thanks to the Matlab wrapper implemented in Caffe. In order to visualize a model, it is necessary to create the number of *<.prototxt>* files corresponding to the number of its layers. After training, and having an appropriate *<.caffemodel>* file with saved weights and biases, it is necessary to call Caffe from Matlab using each of those *<.prototxt>* files as an argument. The received values produced by each layer should be visualized. The visualization example of how Model 4 encodes and decodes the digit "5" is depicted in Fig. 11. The visualizations for other Models look similar. Fig. 11 should be read together with the graphical representation of the Model 4 in Fig. 3 (right). Processing results of intermediate layers are stored in blobs. Blobs are just matrixes, depicted by yellow octagons in Fig. 3 (right) and we visualized their content in Fig. 11. Each panel in Fig. 11 corresponds to the appropriate blob in Fig. 3 (right). The title of each panel in Fig. 11 starts with the same name as the corresponding blob in Fig. 3 (right). As seen in Fig. 3 (right) some blobs have two inputs. For example, the blob "*conv1*" has two inputs: from the convolutional layer *conv1* and from the layer *sig1en* which uses this blob "*conv1*" as an input, performs on that input a transformation by a sigmoid activation function and then stores the result back to the same blob "*conv1*". Therefore in the titles of the appropriate panels in Fig. 11, the title *conv1/conv1* means that this panel visualizes the content of the blob "*conv1*" stored by convolutional layer *conv1* and the title *conv1/sig1en* means that this panel visualizes the content of the blob "*conv1*" stored by sigmoid layer *sig1en*. The same approach is used for all other panels in Fig. 11. The next element of the title of each panel is the size of appropriate blob: for example, *24x24x26* in the panel *conv1/conv1* means that the blob "*conv1*" contains 26 feature maps, each map has size 24x24 elements; *1x1x250* in the panel *ip1encode/sig3en* means that the blob "*ip1encode*" contains 250 neurons with size 1x1 elements; and so on. The titles of all panels (except for fully-connected layers) have square brackets where we indicated the minimum and maximum values of the appropriate blob. The panels which visualize the blobs "*ip1encode*", "*ip2encode*" and "*ip1decode*" belonging to fully-connected layers, contain their numerical representation on the ordinate axis. Thus, the title of every single panel in Fig. 11 clearly describes what processing results are provided by every intermediate layer of a deep network. Such visualizations, and especially a numerical representation of the blobs allowed us to understand that in the failed experiments, the outputs of *deconv2* and *deconv1* layers were negatively or positively saturated, and therefore the pixels of the final reconstructed image in the layer *deconv1neursig* had the value 0 or 1 (obtained by a sigmoid function as a preparation step to calculate a cross-entropy loss, see expression (1)) and the values of loss layers were NaN (Not A Number) during the training.

**Table 6. Quality of reconstruction of random digits '2', '4', '5' and '8' by all unsupervised models**

| Model, Trainable parameters $|w+b|$, Network size (elements) | | | | |
|---|---|---|---|---|
| Model, Trainable parameters \|w+b\|, Network size (elements) | 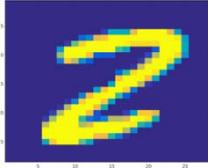 | 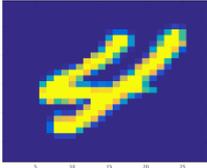 | 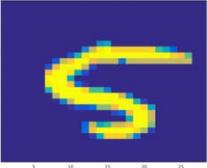 | 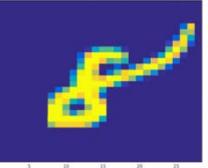 |
| Deep AE, 2823K, 3502 | 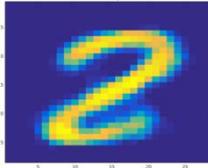 | 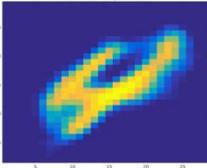 | 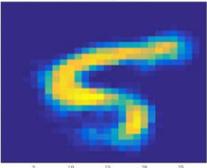 | 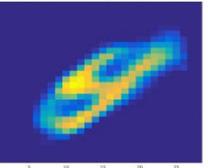 |
| Model 1, 297K, 7990 | 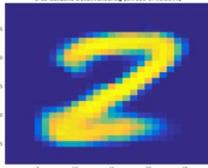 | 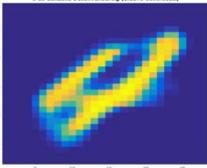 | 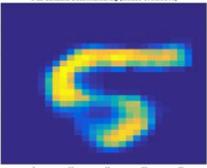 | 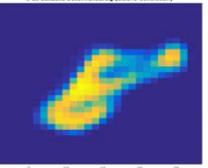 |
| Model 2, 318K, 25814 | 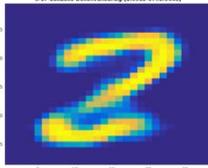 | 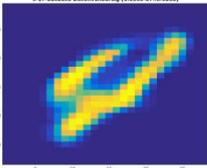 | 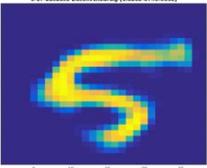 | 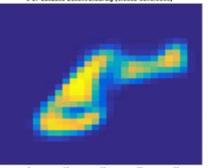 |
| Model 3, 338K, 43702 | 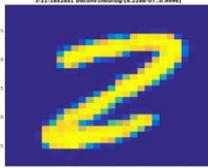 | 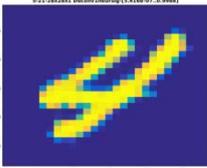 | 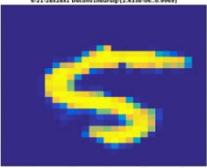 | 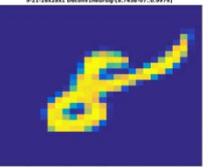 |
| Model 4, 338K, 43702 | 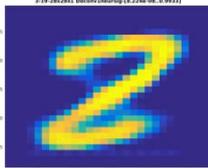 | 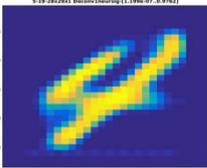 | 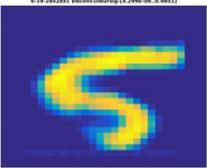 | 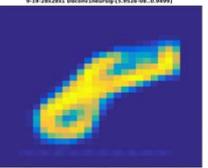 |
| Model 5, 338K, 43702 | 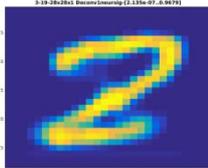 | 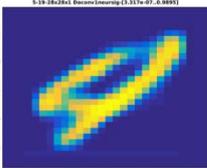 | 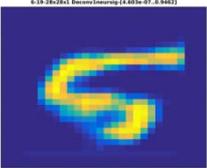 | 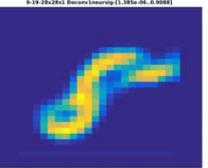 |

All developed Caffe .prototxt files along with Matlab-based visualization scripts are included in supplementary materials and will be made also available in the Caffe user group [58] and on our lab web-page [59]. Note, that the developed Models work in Caffe version used/contributed by Noh et al. [42] (the date of the files in this version is Jun 16, 2015). We ran our models in the latest version we have (the date of files in this version is Apr 05, 2016). It seems, in the newer versions (after Jun 16, 2015) the Caffe developers have changed: (i) the syntax of layer descriptions – from "layers" to "layer", and layers' types from "CONVOLUTION" to "Convolution", etc. and (ii) the internal representation of fully-connected (InnerProduct) layers: it is a 2D array now, not a 4D, as it was in the previous version(s). To deal with these issues it is necessary to change the syntax in the .prototxt files accordingly and to change the dimensionality of the blob of the last fully-connected layer *ip1decode* before the first deconvolution layer *decode2* in the decoder part using the *reshape* layer as follows: *<layer {name: "reshape" type: "Reshape" bottom: "ip1decode" top: "ip1decodesh" reshape_param { shape { dim: 0 dim: 0 dim: 1 dim: 1 }}}>* and use the variable *"ip1decodesh"* as an input to the following layer *decode2*.

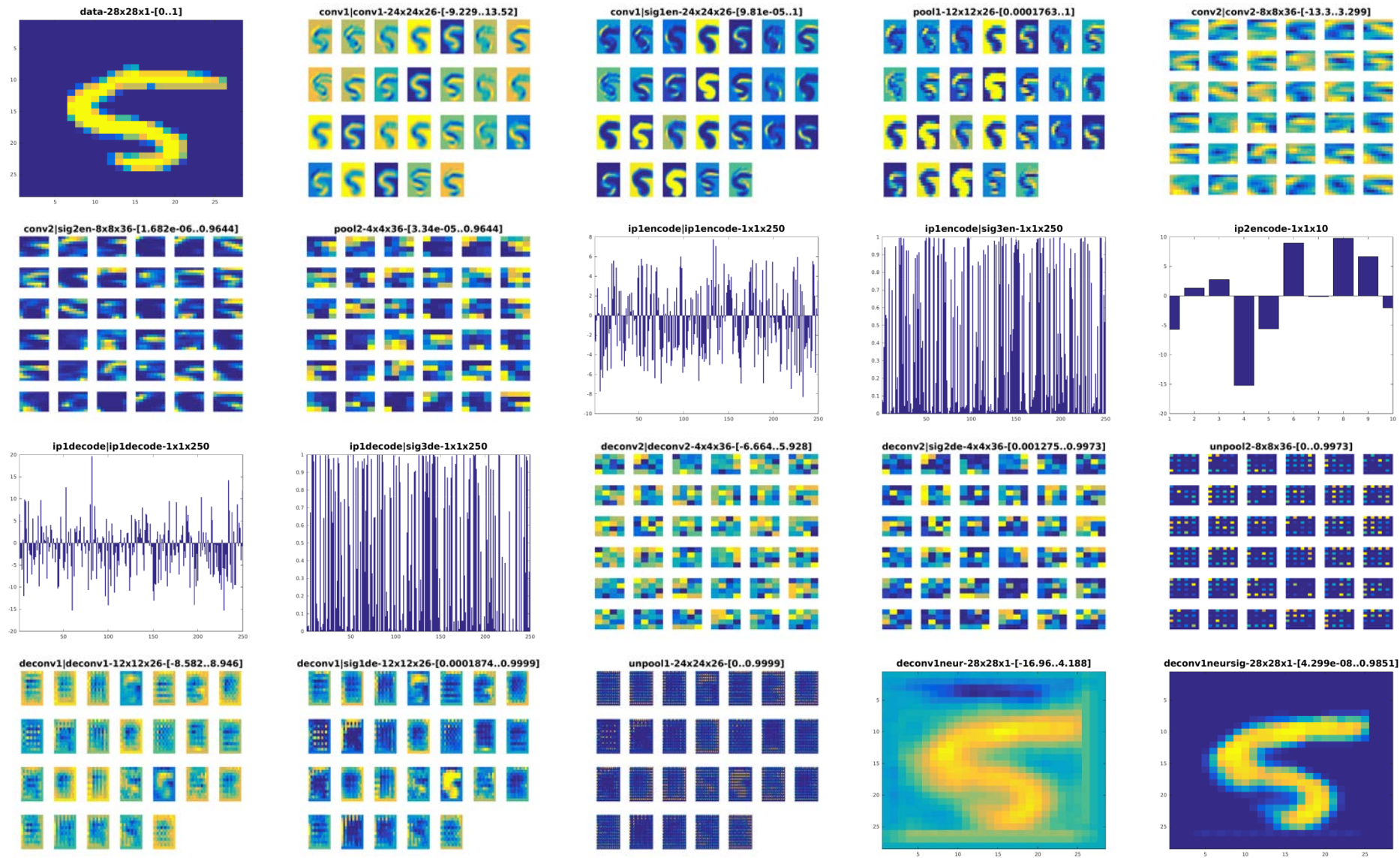

**Fig. 11. Visualization of encoding and decoding digit '5' by 10D Model 4.**
Each panel visualizes the output of appropriate layer of Model 4. The title of each panel describes a type of the layer (e.g. conv1, pool1, conv2, pool2, and so on), a size of the layer (e.g. 24x24x26 means 26 feature maps with 24x24 elements each) and a range of the layer's output [min..max]. See more explanation in the next to last paragraph of Section 4.3.

# 5. Discussion

### 5.1. Why does Model 3 WITH switches provide the worst dimensionality reduction and unsupervised clustering?

Model 3 featured pooling-unpooling layers WITH switches. It reached a much smaller value of our cost function (Table 4) and provided practically an ideal reconstruction of the input images (Table 6). However, it fulfilled the dimensionality reduction (8.75% and 3.75% by 10D and 30D Model 3) and unsupervised clustering tasks (Fig. 8) in the worst way. In this section we discuss this interesting situation.

One explanation is mentioned by Zhao et al. in their paper [27]: *the MNIST dataset is too simple a problem for such a model as Model 3*. Our Model 3 is similar to their SWWAE and we have presented the same perfect reconstruction of MNIST images as in their Figure 2. Zhao et al. stated that "We reason that SWWAE not working so well … is due to the fact that reconstructing MNIST digits is overly easy for SWWAE. … Since MNIST is a roughly binary dataset (0/1) and thus within unpooling stage, decoding *does not necessarily demand* the information from 'what' for reconstruction; i.e., it could get perfect reconstruction by pinning 1 on the positions indicated by 'where'. Therefore, we believe that reconstructing MNIST dataset renders insufficient regularization on the encoding pathway" [27]. In addition to the statement above our results show that a CAE model with pooling-unpooling layers and WITHOUT switches (our Model 4) in its decoder part *does not necessarily demand* the "where" information for successful reconstruction, having the "what" information. Model 4 restores the max-pooled features in the predefined positions [0, 0] (so-called "centralized" unpooling [29, 47]) of the unpooled feature maps and, therefore, prevents too much self-adapting. Such "centralized" unpooling plays a role of a peculiar penalty which penalizes sensitivity of a model to the input and encourages a robustness of the learned representation [19]. Model 4 does not reconstruct an input image perfectly; however, it showed the best quality of dimensionality reduction and unsupervised clustering among all considered unsupervised models.

We present an alternate explanation for Model 3's poor dimensionality reduction and unsupervised clustering performance. The principal difference between Model 3 and other considered unsupervised models is that the use of switch variables considerably increases (inflates) the dimensionality of internal representation of Model 3. Model 3 has the low-dimensional internal code and the switch variables used for reconstruction [15], but all other unsupervised models have the low-dimensional internal code only. Thus Model 3 is not a contractive AE anymore. The dimension of its internal representation is much bigger than the dimension of the input MNIST image (see simple calculations below). Model 3 simply learns the identity function in a trivial way and produces useless, uninteresting features; its internal code is overcomplete [15]. It corresponds to the description of the overcompleteness of features by Norouzi et al. [25] as the following: "What happens is that after a few iterations of parameter update, sampled images become very close to the original ones [which means perfect reconstruction], and the learning signal disappears". We see confirmation that the features in Model 3 are overcomplete in Fig. 8 (right). The last hidden 10D internal code of our Model 3 is so overcomplete, that even the state-of-the-art t-SNE method [55] cannot find any appropriate similarity between the datapoints; it can barely distinguish only two clusters: digit "1" (label "2", orange) and digit "6" (label "7", blue).

As stated in [15, 24-25, 27] the sparsity of an overcomplete code is low. Specifically, Zhao et al. [27] stated in the quotation above that the regularization of their model is insufficient, which means the sparsity is low. Similarly to Zeiler et al. [28], who calculated relative sparsity of feature maps, we have calculated sparsity of the hidden internal code for all our Models (see the last column of Table 4). We have used the sparsity measure [60]

$$\ell_\varepsilon^0 = \#\{\, j, c_j \le \varepsilon \,\}, \tag{2}$$

where $\#$ is a number which describes the sparsity of a vector $\vec{c} = [c_1, c_2, \ldots c_N]$, $N$ is a dimensionality of the analyzed vector $\vec{c}$, $\varepsilon$ is a threshold. Thus, we calculate the number of elements $c_j$ in vector $\vec{c}$ that are smaller than a threshold $\varepsilon$, which means bigger value specifies bigger level of sparsity. We did three experiments for three values of the threshold $\varepsilon$: 10%, 20% and 30% of the maximum value in each analyzed internal code for all models presented in Table 4. We have received similar results for all thresholds $\varepsilon$ and we have specified the average value of sparsity for 30% threshold for all three runs in Table 4. Rather than specify the sparsity $\ell_\varepsilon^0$ as a number (which actually will be hard to interpret and to compare between all 2D, 10D and 30D models), we specified it as a percentage of the elements of the internal code which are smaller than a threshold $\varepsilon$ (30%) for the appropriate internal code. For example, the sparsity 70.9% for 2D Model 4 (Table 4) means that 70.9% of total elements of that code within all 10000 test patterns (i.e. 2D * 10000 = 20000 total elements) are less than 30% of the maximum value found in these 20000 total elements. Such a relative measure of $\ell_\varepsilon^0$ allows us to compare all models. Note, that 1% here means 200 elements for 2D internal code, 1000 elements for 10D internal code and 3000 elements for 30D internal code. The analysis of Table 4 has shown that the sparsity of the internal code of the Model 3 is *less* than all other models: in the 2D case it is 30.8% against the interval of a minimum 59.3% in a Siamese network and a maximum 79.7% in Model 2, in the 10D case it is 67.4% against the interval of a minimum 68.9% in Model 5 and a maximum 75.8% in Model 2, and in the 30D case it is 74.8% against the interval of a minimum 77.6% in Model 5 and a maximum 86.9% in Model 4. Our results show no clear relation between better sparsity of the internal code and better average per-class classification error: we just state that obviously the sparsity of the internal code of the Model 3 is *less* than other successfully-working models.

The well-known regularization techniques, for example dropout [30] and weight decay [61], aim to improve the sparsity of deep models and prevent trainable parameters from growing too large. Therefore we have estimated the sparsity of the trainable parameters of all researched models using expression (2) and the approach described above. Again, for the 30% threshold $\varepsilon$, the successfully-working models have shown bigger sparsity (12.3-12.4%), while Model 3 has shown a lower value of 11.8%.

The overcompleteness of Model 3's internal representation makes reconstruction "overly easy" [27], explaining its good reconstruction but poor clustering performance. A complementary explanation arises when we realize that the essential function of an AE is compression: finding a low-dimensional code to represent a high-dimensional input [5]. Such a low-dimensional but information-rich code represents a useful clustering of the data and can facilitate effective classification. We propose that switch variables undermine the autoencoder's principal role as a compressor. To illustrate this, consider the compression ratio achieved by Model 3. One simple way to approximate this compression ratio is to assume that switch variables and other values are uniformly distributed over their ranges. Under this simple assumption, a MNIST image of 28 x 28 pixels in the range [0,255] requires 784 bytes to encode, while the 2D, 10D, and 30D single-precision codes require 8, 40, and 120 bytes respectively. Each of the 3744 switch variables stored by Model 3's first pooling layer is an integer address in the range [0,575], and thus requires at least 9 bits to encode. The 576 switch variables in the second pooling layer take values in the range [0,63], and require 6 bits to encode. Thus Model 3's switch variables and internal code together total between 4652 and 4764 bytes of information – significantly more information than was stored in the input image itself, and giving a negative compression ratio of -490%. The use of switch variables causes Model 3 to inflate rather than compress the input data. In contrast, all other unsupervised models, which produce only the 2D, 10D, or 30D internal codes, achieve compression ratios of 99%, 95%, and 85% respectively. A more thorough analysis of compression ratio could measure self-information or entropy, but we suppose that this simple calculation is sufficiently illustrative.

Model 3's inflation – rather than compression – of the input data short-circuits the AE training process, which normally guides the AE to achieve minimal reconstruction error through effective compression. The extra information cached in switch variables allow Model 3 to achieve good reconstruction during training without learning an effective compression scheme, which explains why the resulting 2D, 10D, and 30D internal codes have low clustering and classification value.

### *5.2. Experiments with ReLU activation function*

A ReLU activation function has been proven better for supervised deep machine learning tasks in a lot of research studies because it provides better sparsification of a deep NN model, and models with sparser coding provide better classification results. This useful property of a ReLU function should serve for unsupervised models too. The summary on the use of different activation functions in the existing solutions of CAE are the following: [21, 24-25, 35] have used sigmoid (logistic), [15] has used sparsifying logistic, [26] has used scaled hyperbolic tangent, [15, 39] have used hyperbolic tangent and [20, 23, 27, 40-41] have used ReLU.

When we have used a ReLU function in our Models they, surprisingly, did not train. When we ran the experiments using the same initialization parameters for weights and biases as it was described in Section 4.2 above, i.e. *<bias_filler {type: "constant"}>* for all layers, *<weight_filler {type: "xavier"}>* for conv/deconv layers, *<weight_filler {type: "gaussian" std: 1 sparse: 25}>* for fully-connected layers, the Models contained the value NaN (Not A Number) in all feature maps and the values of our cost function during training were NaN too. When we have initialized the weights and biases within the smaller interval, i.e. we have used *<weight_filler {type: "gaussian" std: 0.01}>*, *<bias_filler {type: "constant" value: 0}>* for all layers, the Models calculated the values of our cost function in the first training iteration only. After the first iteration both values were set to 307.24; 37.88 and further they changed only slightly; staying more-less the same until training stopped. In our opinion, it may be necessary to find some correct initialization of a CAE model with a ReLU activation function or use some ideas described in the PyLearn user group [62]. Then, hopefully, due to the better sparsification properties, a ReLU activation function could be able to provide better results than our Models with sigmoid and hyperbolic tangent activation functions. This will be one of the future lines of our research.

## 6. Conclusions

The development of several deep convolutional auto-encoder models in the Caffe deep learning framework and their experimental evaluation on the MNIST dataset are presented in this paper. In contrast to the deep fully-connected auto-encoder proposed by Hinton et al. [5], convolutional auto-encoders allow using the desirable properties of convolutional neural networks for image processing tasks while working within an unsupervised learning paradigm.

We have created and researched five convolutional auto-encoder models in Caffe: (i) Model 1 which contains two convolutional layers followed by two fully-connected layers in the encoder part and, inversely, one fully-connected layer followed by two deconvolution layers in the decoder part, (ii) Model 2 which contains two pairs of convolutional and pooling layers followed by two fully-connected layers in the encoder part and, inversely, one fully-connected layer followed by only two deconvolution layers in the decoder part, (iii) Model 3 which contains two pairs of convolutional and pooling layers followed by two fully-connected layers in the encoder part and, inversely, one fully-connected layer followed by two pairs of deconvolution and unpooling layers WITH the use of switch variables in the decoder part, (iv)

Model 4 which is the same as Model 3 but WITHOUT the use of switch variables in the decoder part and (v) Model 5 which is the same as Model 4 except of using a hyperbolic tangent activation function in all layers (instead of sigmoid in Models 1-4). The hidden low-dimensional internal code learned by these Models in an unsupervised way was used as an input for a linear classifier and a standard one-hidden-layer perceptron and the classification errors for each Model were estimated.

Our results show that the developed Models 1-2 and 4-5 provide very good results of the dimensionality reduction and unsupervised clustering tasks and small classification errors. Specifically, Model 1 and Model 2 without pooling – unpooling layers provide slightly better results (2.65% by 30D Model 1 and 2.54% by 30D Model 2) than a deep fully-connected auto-encoder (2.85% by 30D deep auto-encoder). *Model 4 with pooling-unpooling layers and WITHOUT switches shows the best result (2.19% by 30D Model 4).* Model 5 shows that the use of a hyperbolic tangent activation function instead sigmoid provides worse results (3.36% by 10D Model 5). Model 3 with pooling-unpooling layers and WITH switches shows a practically an ideal reconstruction of the input images, but the worst fulfilment of the dimensionality reduction and unsupervised clustering tasks and therefore large classification errors (8.75% and 3.75% by 10D and 30D Model 3). We think this is because Model 3's use of switch variables considerably inflates the dimensionality of its internal representation. Model 3 most likely learns only the trivial identity mapping.

During the creation of Models of a convolutional auto-encoder we have followed several rules of thumb, mentioned in Section 3 above, which are used by many machine learning researchers every day. The paper also discusses practical details of the creation of a deep convolutional auto-encoder in the very popular Caffe deep learning framework. All developed Caffe *.prototxt* files along with Matlab-based visualization scripts are included in supplementary materials and will be made also available in the Caffe user group [58] and on our lab web-page [59]. We believe that our approach and results, presented in this paper, could help other researchers to build efficient deep neural network architectures in future.

The application of the developed Models of a deep convolutional auto-encoder for the analysis of input images in the neuroscience field is the direction of our future research.

## Acknowledgements

We would like to thank the Caffe developers (the Berkeley Vision and Learning Center, UC Berkeley) for creating such a powerful framework for deep machine learning research. We thank Karim Ali (CCBN) for help with Caffe installation on *Hodgkin* and *Polaris1*, Hyeonwoo Noh (POSTECH, Korea) for using his Caffe implementation of the unpooling layer and discussions on some results presented in this paper, and Dr. Robert Sutherland (CCBN) for help with financial support.

## Bios

**Volodymyr Turchenko** received a M.Sc. in System Engineering (honors) from Brest Polytechnic University, Belarus and a Ph.D. in Computer Engineering from Lviv Polytechnic National University, Ukraine. He worked as a postdoctoral fellow funded by the FP7 Marie Curie IIF grant at the University of Calabria, Italy and as a Fulbright scholar at the University of Tennessee researching parallelization schemes of artificial neural networks. He is currently a postdoctoral fellow at the Canadian Centre for Behavioural Neuroscience at the University of Lethbridge. His research interests include theory and application of artificial neural networks and deep machine learning.

**Eric Chalmers** received a B.Sc. in Electrical Engineering in 2011, and a Ph.D. in Electrical & Computer Engineering in 2015 from the University of Alberta, Canada. He is currently a postdoctoral fellow at the Canadian Centre for Behavioural Neuroscience at the University of Lethbridge. His research interests include brain-inspired machine learning and intelligent systems.

**Artur Luczak** received a M.Sc. in Biomedical Engineering from Wroclaw University of Technology, and Ph.D. from Jagiellonian University, Poland. As a postdoctoral fellow he worked at Yale and Rutgers University studying neural information processing. Currently he is a faculty at the Canadian Centre for Behavioural Neuroscience at the University of Lethbridge, where his research involves high density neuron recordings to study brain dynamics.